%% file: main.tex
\documentclass[runningheads]{llncs}
\usepackage[T1]{fontenc}

\usepackage{amsmath,graphicx,hyperref}
\usepackage{amsmath,amssymb}
\usepackage{bm}
\usepackage{xcolor}
\usepackage{colortbl}
\usepackage{booktabs}
\usepackage{wrapfig}
\usepackage{multirow}
\usepackage{url}

\def\BibTeX{{\rm B\kern-.05em{\sc i\kern-.025em b}\kern-.08emT\kern-.1667em\lower.7ex\hbox{E}\kern-.125emX}}

\def\etal{\emph{et al.}}

\title{Polynomial Dice Loss\\ for Medical Image Segmentation}
\author{Hiroaki Aizawa}
\institute{Graduate School of Advanced Science and Engineering\\
Hiroshima University\\
Higashi-Hiroshima, Japan\\
\email{hiroaki-aizawa@hiroshima-u.ac.jp}}

\input{tabs}

\input{figs}

\begin{document}
%
\maketitle
\begin{abstract}
Medical image segmentation is a fundamental task for medical image processing and computer-assisted intervention, yet data imbalance and small lesion detection pose significant challenges. Dice Loss, which measures the overlap between predicted and ground truth regions, is widely used to mitigate these issues. To further emphasize its properties, we propose Polynomial Dice Loss, a polynomial extension of Dice Loss. Specifically, by leveraging the geometric characteristics of Dice Loss and formulating the loss function as a polynomial representation via Taylor expansion, we enable the adjustment of the contribution of higher-order components to the loss function. In our experiments, we evaluate the proposed method against loss functions derived from conventional Dice and Tversky coefficients. Experimental results and further analysis show that the polynomial formulation provides a simple way to control the loss shape and achieves competitive performance across multiple segmentation settings.
\end{abstract}

\keywords{Medical Image Segmentation \and Dice Loss \and Polynomial Loss Function \and Taylor expansion}


\section{Introduction}
Medical image segmentation is a fundamental component of medical image analysis and computer-assisted intervention, supporting tasks such as disease screening, treatment planning, and longitudinal follow-up. In many clinical scenarios, however, the target structures occupy only a small fraction of the image. This is common in lesion segmentation, early-stage findings, and organ sub-structures. Such settings impose heavy class imbalance and make learning highly sensitive to small localization errors, where a few misclassified pixels can dominate clinical utility.

Beyond imbalance, medical datasets often exhibit limited annotation budgets, heterogeneous imaging protocols, and noisy labels due to inter-observer variability. These factors can amplify optimization instability and degrade generalization, especially when the foreground is tiny. As a result, the design of the loss function becomes a key lever for stable training and reliable detection of small structures, often as important as architecture choices or data augmentation.

Dice loss~\cite{diceloss} is a common choice because it optimizes overlap directly and often detects small structures better than cross-entropy~\cite{ce}. Many variants of Dice and Tversky losses report strong results in CT and MRI segmentation~\cite{dice+ce,dice+focal,noisedice,adaptivedice,generaliseddiceloss}. Dice Loss is built on the Dice Index, a similarity measure that quantifies overlap between prediction and ground truth and naturally emphasizes small overlaps, such as those in minority classes, more than Cross-Entropy Loss. Optimizing Dice Loss increases the contribution of minority-class errors to the average loss, improving detection under imbalance. Generalizations based on the Tversky Index and their extensions have also been proposed~\cite{tverskyloss,focaltverskyloss} and are effective for imbalanced data.

The choice and shape of the loss also influence generalization. Cross-Entropy Loss~\cite{ce} is widely used in classification and segmentation, yet it struggles under class imbalance and label noise. To overcome these issues, a complementary direction is to employ polynomial approximations of loss functions~\cite{polyloss,taylorce,tayloroe}. TaylorCE~\cite{taylorce} approximates cross-entropy with a Taylor series, and PolyLoss~\cite{polyloss} adds polynomial coefficients to the expanded form. By tuning these coefficients, PolyLoss recovers Cross-Entropy and Focal Loss~\cite{focalloss} as special cases and adjusts gradient magnitudes to emphasize hard or minority examples. Thus, polynomial parameterizations provide explicit control over a loss function's shape and can improve robustness and generalization.

\overview

Based on these findings, we propose \emph{Polynomial Dice Loss}, a polynomial reformulation of Dice Loss that introduces controllable higher-order components while preserving the overlap-driven objective. We begin with a geometric view of the Dice coefficient and decompose Dice Loss into a scale term and a directional term given by cosine similarity. Approximating the directional term with a Taylor series yields a polynomial form of Dice Loss. Following PolyLoss~\cite{polyloss} and TaylorCE~\cite{taylorce}, we introduce high-order truncation and coefficient controls to shape the loss, enabling more flexible training, as illustrated in Figure~\ref{fig:overview}. In experiments, we evaluate the proposed losses alongside methods derived from the Dice and Tversky indices. The results and analyses indicate that the polynomial formulation improves controllability of Dice-based losses and yields competitive or improved performance in several settings.

\noindent\textbf{Contributions.}
\begin{itemize}
  \item \textbf{Polynomial Dice Loss via Taylor expansion.}
  By applying a Taylor expansion to the directional term, we derive a polynomial representation of Dice Loss and introduce explicit controls through truncation order and coefficient weighting, enabling systematic shaping of the loss landscape.

  \item \textbf{Empirical validation and analysis.}
  We evaluate the proposed losses against standard Dice and Tversky-family baselines and provide analyses on how polynomial orders and coefficients affect training behavior and empirical performance, particularly under class imbalance and tiny foreground regions.
\end{itemize}

\section{Methodology}
\label{sec:method}
\subsection{Preliminaries: Dice Loss}
Let $\boldsymbol{y}\in\{0,1\}^{n}$ be the ground-truth mask and $\hat{\boldsymbol{y}}\in\{0,1\}^{n}$ be the predicted mask, where $n=H \times W$ after flattening an image. In practice, $\hat{\boldsymbol{y}}$ is often a \emph{soft} prediction (e.g., a sigmoid/softmax output), i.e., $\hat{\boldsymbol{y}}\in[0,1]^n$, but we keep the same notation for simplicity. Throughout this paper, $\langle\cdot,\cdot\rangle$ denotes the dot product and $\|\cdot\|_2$ the Euclidean norm. 

Dice Loss is derived from the Dice coefficient, which measures the overlap between two sets (or masks). Interpreting $\langle \boldsymbol{y},\hat{\boldsymbol{y}}\rangle$ as the (soft) intersection and $\|\boldsymbol{y}\|_2^2+\|\hat{\boldsymbol{y}}\|_2^2$ as the (soft) sum of volumes, Dice Loss is defined as
\begin{align}
\ell_{\text{Dice}}(\boldsymbol{y},\hat{\boldsymbol{y}})
= 1 - \frac{2\,\langle \boldsymbol{y},\hat{\boldsymbol{y}}\rangle + \delta}{\|\boldsymbol{y}\|_{2}^{2}+\|\hat{\boldsymbol{y}}\|_{2}^{2}+\delta},
\end{align}
where $\delta>0$ is a small constant for numerical stability. The smoothing term prevents division by zero and stabilizes optimization when the foreground is extremely small or absent (e.g., $\|\boldsymbol{y}\|_2^2 \approx 0$ in negative samples). Unless otherwise noted, our subsequent formulation is presented with $\delta=0$ only for analytic clarity; the actual implementation uses $\delta>0$.

\paragraph{Geometric decomposition.}
For geometric insight~\cite{adaptivedice}, set $\delta=0$ and rewrite Dice Loss as
\begin{align}
\ell_{\text{Dice}}
&= 1 - \frac{2\,\langle \boldsymbol{y},\hat{\boldsymbol{y}}\rangle}{\|\boldsymbol{y}\|_{2}^{2}+\|\hat{\boldsymbol{y}}\|_{2}^{2}} \notag \\
&= 1 - \underbrace{\frac{2\|\boldsymbol{y}\|_{2}\,\|\hat{\boldsymbol{y}}\|_{2}}{\|\boldsymbol{y}\|_{2}^{2}+\|\hat{\boldsymbol{y}}\|_{2}^{2}}}_{\text{scale:} s}\,
\underbrace{\frac{\langle \boldsymbol{y},\hat{\boldsymbol{y}}\rangle}{\|\boldsymbol{y}\|_{2}\,\|\hat{\boldsymbol{y}}\|_{2}}}_{\text{direction:} \cos\theta}.
\label{eq:geometric}
\end{align}
Here, $\theta$ is the angle between the two vectors. This decomposition makes explicit that Dice Loss depends on \emph{(i) a magnitude agreement} and \emph{(ii) an alignment agreement}:
\begin{itemize}
\item \textbf{Scale term $s$.}
The factor $s \in (0,1]$ compares the magnitudes $\|\boldsymbol{y}\|_2$ and $\|\hat{\boldsymbol{y}}\|_2$.
Using the arithmetic-geometric mean inequality,
$\|\boldsymbol{y}\|_2^2 + \|\hat{\boldsymbol{y}}\|_2^2 \ge 2\|\boldsymbol{y}\|_2\|\hat{\boldsymbol{y}}\|_2$,
we have $s \le 1$, and $s=1$ holds if and only if $\|\boldsymbol{y}\|_2=\|\hat{\boldsymbol{y}}\|_2$.
Thus, $s$ penalizes \emph{volume mismatch} between prediction and ground truth.

\item \textbf{Directional term $\cos\theta$.}
The factor $\cos\theta = \frac{\langle \boldsymbol{y},\hat{\boldsymbol{y}}\rangle}{\|\boldsymbol{y}\|_{2}\|\hat{\boldsymbol{y}}\|_{2}}$ is the cosine similarity and quantifies how well the predicted mask aligns with the ground-truth mask in direction.
It equals $1$ when $\hat{\boldsymbol{y}}$ is perfectly aligned with $\boldsymbol{y}$ (up to a positive scaling), and decreases as the prediction deviates.
\end{itemize}

\paragraph{Normalized view.}
If both vectors are $L_{2}$-normalized, i.e., $\|\boldsymbol{y}\|_2=\|\hat{\boldsymbol{y}}\|_2=1$, then $s=1$ and the loss reduces to a purely angular form,
\begin{align}
\ell_{\text{Dice}} = 1-\cos\theta.
\end{align}
This normalized perspective is particularly useful for our method because it isolates the \emph{directional component} as a smooth similarity term, which can be approximated by a polynomial expansion in later sections. In contrast, the scale term primarily controls the foreground volume, while the directional term captures the spatial agreement that we aim to reshape via our polynomial formulation.

\subsection{Polynomial Dice Loss}
Based on the geometric interpretation (Eq.~\eqref{eq:geometric}), we derive a polynomial form of Dice Loss by approximating its angular component. Recall that Dice Loss can be written as
\begin{align}
\ell_{\mathrm{Dice}} = 1 - s \cos\theta,
\end{align}
where $s$ is the scale term and $\theta$ is the angle between the prediction and the target in the flattened vector space. The key benefit of this form is that it separates (i) \emph{volume agreement} through $s$ and (ii) \emph{alignment agreement} through $\cos\theta$. Our polynomial formulation targets the alignment part.

\paragraph{Taylor approximation of the directional term.}
During training, the prediction $\hat{\boldsymbol{y}}$ gradually approaches the target $\boldsymbol{y}$, so $\theta$ tends to be small in well-optimized regimes. This motivates approximating $\cos\theta$ using a Taylor expansion around $\theta=0$:
\begin{align}
\cos\theta
= 1 - \frac{\theta^2}{2!} + \frac{\theta^4}{4!} - \cdots.
\end{align}
Assuming the scale term $s$ is treated as fixed for this derivation, substituting the expansion into $\ell_{\mathrm{Dice}}$ yields
\begin{align}
\ell_{\text{Dice}}
&\approx 1 - s \left(1 - \frac{\theta^2}{2!} + \frac{\theta^4}{4!} - \cdots\right) \notag\\
&= (1 - s) + s \sum_{k=1}^{\infty} \frac{(-1)^{k-1}}{(2k)!}\,\theta^{2k}.
\end{align}
Defining the series coefficients $\alpha_k = \frac{(-1)^{k-1}}{(2k)!}$, we obtain the \emph{Polynomial Dice Loss}:
\begin{align}
\ell_{\text{PolyDice}} = (1 - s) + s \sum_{k=1}^{\infty} \alpha_k\,\theta^{2k}.
\end{align}

\paragraph{Why this view is useful.}
The purpose of this polynomial expression is not simply to ``model high-order effects.'' Instead, it exposes Dice Loss as an infinite sum of even-order angular penalties, which gives us a simple handle to \emph{regulate how strongly misalignment is penalized}. In particular, the higher-order terms $\theta^{2k}$ dominate when $\theta$ is large (failed predictions). If we ignore some of these terms, the penalty profile in the high-error region changes, which lets us make the loss harsher or more conservative for incorrect predictions while keeping the Dice structure.

Importantly, the truncated polynomial is not intended to be a uniformly accurate approximation of $\cos\theta$ over the entire interval $\theta \in [0,\pi]$. Rather, we use the Taylor expansion as a parameterized surrogate that preserves the local Dice behavior near well-aligned predictions while deliberately reshaping the tail behavior for poorly aligned predictions.

\paragraph{Computing the angle.}
We compute $\theta$ from the cosine similarity between $\boldsymbol{y}$ and $\hat{\boldsymbol{y}}$:
\begin{align}
\theta = \arccos\!\left(
\frac{\langle \boldsymbol{y}, \hat{\boldsymbol{y}} \rangle}{\|\boldsymbol{y}\|_{2}\,\|\hat{\boldsymbol{y}}\|_{2}}
\right).
\end{align}

For multi-class segmentation, we compute the loss independently for each class channel using one-hot ground-truth masks and softmax predictions, and then average the class-wise losses. This class-wise aggregation is used for Dice Loss, DropDice, and PolyDice-1 in all multi-class experiments. The background channel is included in the same manner as the foreground classes.

\visdropandpoly

\paragraph{DropDice Loss.}
Following PolyLoss~\cite{polyloss}, we examine variants of Polynomial Dice Loss parameterized by the polynomial order. We define a simplified version by truncating the Taylor expansion at a fixed order $K$, which yields the DropDice Loss:
\begin{align}
\ell_{\text{DropDice}}^{(K)} = (1-s) + s \sum_{k=1}^K \alpha_k \theta^{2k}.
\label{eq:dropdice}
\end{align}

This truncation balances expressiveness and computational cost. Discarding higher-order terms prioritizes lower-order components that capture coarse alignment errors and may improve training stability. In practice, we terminate the sum at a fixed $K$ such as $1$, $2$, or $3$, depending on the experimental setting.

\paragraph{PolyDice-1 Loss.}
Although $\ell_{\text{DropDice}}^{(K)}$ introduces a controllable order $K$, tuning multiple coefficients in addition to $K$ is impractical. Leng~\etal~\cite{polyloss} showed that modifying only the leading term in a polynomial expansion often provides strong gains with minimal tuning effort. Following this finding, we simplify the formulation by adjusting only the first term:
\begin{align}
\ell_{\text{PolyDice-1}} = (1-s) + s\,[\alpha_1 + \epsilon_1]\,\theta^2,
\label{eq:poly1dice}
\end{align}
where $\epsilon_1$ is a tunable parameter. This introduces a single extra degree of freedom that directly scales the dominant angular penalty component. Larger $\epsilon_1$ increases the penalty and gradient magnitude in high-error regions, encouraging stronger correction, while smaller or negative $\epsilon_1$ yields a more conservative penalty. Because it introduces only one additional coefficient, we use $\ell_{\text{PolyDice-1}}$ as the primary practical variant in our experiments. We further illustrate how $K$ and $\epsilon_1$ regulate the penalty strength as functions of $\theta$ in Sec.~\ref{sec:intuition}.

\subsection{Intuitive Understanding of Polynomial Dice Loss}
\label{sec:intuition}
To clarify the roles of our hyperparameters, we visualize $\ell_{\text{DropDice}}^{(K)}$ and $\ell_{\text{PolyDice-1}}$ as functions of the angular error $\theta \in [0,\pi]$ in Figures~\ref{fig:dropdice} and~\ref{fig:poly1dice}. In all plots, we fix the scale term to $s=0.1$ to isolate the effect of the \emph{directional} component. Here, $\theta \approx 0$ corresponds to a well-aligned prediction, whereas large $\theta$ indicates a failed prediction with low alignment.

\paragraph{DropDice Loss: truncation controls stronger/weaker penalties for failures.}
For $\ell_{\text{DropDice}}^{(K)}$ in Eq.~\eqref{eq:dropdice}, we vary the truncation order $K$. The curves are nearly identical near $\theta=0$ but differ substantially for large $\theta$. This is expected because the polynomial view exposes Dice Loss as a sum of even-order terms, and these terms dominate the behavior when $\theta$ is large.

Compared with Dice Loss, the difference appears mainly in the high-error region. With smaller $K$ (e.g., $K=1$), the truncated form assigns \emph{stronger penalties} to large $\theta$, which mitigates under-penalization when predictions fail. As $K$ increases, the curve approaches the original Dice formulation, indicating that $K$ regulates how quickly the penalty saturates for large errors. In other words, $K$ controls the \emph{tail} behavior: smaller $K$ yields stronger penalties for failures, while larger $K$ recovers a more conservative penalty profile closer to standard Dice Loss.

\paragraph{PolyDice-1 Loss: a single coefficient adjusts penalty strength.}
For $\ell_{\text{PolyDice-1}}$ in Eq.~\eqref{eq:poly1dice}, we again fix $s=0.1$ and vary $\epsilon_1$. Since PolyDice-1 modifies only the leading term, $\epsilon_1$ directly adjusts the strength of the dominant angular penalty. Relative to Dice Loss, larger $\epsilon_1$ yields \emph{stronger penalties} (and thus larger gradients) in the high-error region, enabling stronger correction when predictions deviate from the target. Conversely, smaller or negative $\epsilon_1$ yields \emph{weaker penalties} for large $\theta$, resulting in a more conservative update.

Overall, these visualizations highlight the main point of our polynomial view: it provides a simple and controllable way to assign stronger or weaker penalties to misaligned predictions, especially in failure cases. This explicit control can improve optimization behavior and training stability in imbalanced medical image segmentation.

\resultseg
\tableunet
\tabletransunet

\section{Evaluation}
\label{sec:evaluation}

\subsection{Evaluation Settings}
\paragraph{Dataset.} We use four public benchmarks: CVC-ClinicDB~\cite{cvc} and Kvasir-SEG~\cite{kvasir} for binary segmentation, and ACDC~\cite{acdc} and Synapse\footnote{\url{https://www.synapse.org/\#!Synapse:syn3193805/wiki/217789}} for multi-class segmentation. Following prior work~\cite{adaptivedice}, since ACDC and Synapse are 3D volumetric datasets, we convert them to 2D by extracting axial slices. Each dataset is split and preprocessed by resizing images to $224\times224$, with data augmentation using \texttt{RandomRotation} and \texttt{RandomHorizontalFlip}.

\paragraph{Comparison Methods.} We compare with Dice Loss (Dice)~\cite{diceloss}, Cross-Entropy Loss (CE)~\cite{ce}, PolyCE-1 Loss~\cite{polyloss}, Tversky Loss (Tversky)~\cite{tverskyloss}, and Focal Tversky Loss (Focal Tversky)~\cite{focaltverskyloss}. For our proposed losses, we select $\epsilon$ by grid search over $[-0.3,-0.2,\dots,0.5]$ using the validation set. We exclude the validation-adaptive loss in~\cite{adaptivedice} from the comparison because it updates the hyperparameter on the validation set during training in a manner similar to bilevel optimization.

\paragraph{Training Details.} We adopt UNet~\cite{unet} and TransUNet~\cite{transunet}. Both models were implemented as 2D convolutional architectures, suitable for the 2D slice-based setting described above. The UNet encoder is randomly initialized, and the TransUNet encoder uses ImageNet-pretrained weights. We used Momentum SGD with a learning rate of 0.01, momentum of 0.9, and weight decay of 0.0001. All models were trained for 200 epochs with a batch size of 24. We performed 5-fold cross-validation and report the mean and standard deviation of the Dice score.

\expdropdice
\exppolydice

\subsection{Performance Comparison}

We discuss the effectiveness of the proposed losses across architectures and datasets. Tables~\ref{tab:unet_results} and~\ref{tab:trans_unet_results} report UNet and TransUNet results, respectively. Figure~\ref{fig:results_seg} provides qualitative examples on Synapse with TransUNet.

From Table~\ref{tab:unet_results}, the proposed losses achieve competitive performance on UNet, with larger gains on multi-class tasks (ACDC and Synapse). This suggests that the polynomial formulation can be beneficial in some Dice-based segmentation settings. On CVC-ClinicDB, Focal Tversky slightly outperforms PolyDice-1, although PolyDice-1 remains competitive.

From Table~\ref{tab:trans_unet_results}, both DropDice and PolyDice-1 performed well on TransUNet for binary and multi-class segmentation. Tversky-based losses were sometimes competitive on binary tasks, but their accuracy dropped on multi-class data. On Synapse, Tversky variants stay around 50--60\%, whereas DropDice and PolyDice-1 reach about 79\%. This indicates that Tversky losses can be effective when foreground regions are few, but their performance is less stable in our multi-class experiments.

By contrast, our polynomial formulation maintains strong performance across models and datasets. It provides flexible control of the penalty by shaping gradients for misclassified predictions (Figure~\ref{fig:poly1dice}), which aids multi-class learning. The tuning cost is small because only one coefficient is added, although the best value remains dataset dependent.

\subsection{Effect of the Order $K$ in $\ell_{\text{DropDice}}^{(K)}$}
\label{sec:exp_drop_dice_k}

Next, we examine the effect of the order $K$ in DropDice. We set $K\in\{1,2,3,10\}$ in Eq.~\eqref{eq:dropdice} and trained UNet on each dataset. Figure~\ref{fig:exp_drop_dice_k} shows the relationship between $K$ and Dice scores. These results indicate that performance varies with $K$ and is dataset dependent. For CVC-ClinicDB and Synapse, larger $K$ yields accuracy comparable to or better than Dice, consistent with $K=10$ approaching the infinite-series limit. Conversely, on ACDC even $K=1$ sometimes surpasses Dice, indicating that adding higher-order terms does not guarantee gains. Selecting an optimal $K$ is therefore nontrivial.

\subsection{Effect of the Coefficient $\epsilon$ in $\ell_{\text{PolyDice-1}}$}

Finally, we examine the effect of the coefficient $\epsilon$ in PolyDice-1 Loss. We use DropDice with $K=1$ as the baseline and vary $\epsilon$ from $-0.3$ to $0.5$ to compare performance. Figure~\ref{fig:exp_poly_dice_epsilon} shows that, as with the truncation order $K$ in DropDice, the best $\epsilon$ is strongly dataset dependent, and validation-based tuning within an appropriate range can improve empirical performance. Thus, while PolyDice-1 Loss controls the loss shape with a single hyperparameter, in practice $\epsilon$ should be selected for each dataset. Nonetheless, PolyDice-1 Loss has a single parameter to tune, making it easy to implement and train in terms of cost, and this tuning cost is modest compared with variants that require multiple coefficients.

\section{Conclusion}
\label{sec:conclusion}
We reformulated Dice Loss by applying a Taylor expansion to its geometric form and introduced DropDice Loss and PolyDice-1 Loss. Across standard models and datasets, performance depended on the truncation order $K$, and PolyDice-1, with a single tunable coefficient, effectively shaped the loss and improved results, especially in multi-class and imbalanced settings with small foregrounds. These findings show that polynomial truncation and coefficient adjustment provide a simple handle to control optimization.

A limitation of this study is that the 3D volumetric datasets are evaluated in a 2D slice-based setting. Evaluating Polynomial Dice Loss with native 3D architectures, such as 3D U-Net, is left for future work.

\bibliographystyle{splncs04}
\bibliography{main}

\end{document}

%% file: tabs.tex
\def\tableunet{
\begin{table*}[t!]
\centering
\caption{Comparison of Dice Similarity Coefficient (\%) for UNet. Results are reported as the mean and standard deviation over 5-fold cross-validation, and higher values indicate better segmentation performance. The best score for each dataset is shown in bold, and shaded rows denote the proposed losses.}
\label{tab:unet_results}
\begin{tabular}{lcccc}
\toprule[0.4mm]
\textbf{Method} & \textbf{CVC-ClinicDB} & \textbf{Kvasir-SEG} & \textbf{ACDC} & \textbf{Synapse} \\
\midrule
Dice Loss & 
$79.31_{\pm 4.69}$ & 
$90.27_{\pm 0.37}$ & 
$91.95_{\pm 2.41}$ & 
$70.44_{\pm 5.47}$ \\

Cross Entropy Loss & 
$77.01_{\pm 3.72}$ & 
$90.14_{\pm 0.44}$ & 
$90.65_{\pm 2.55}$ & 
$63.77_{\pm 6.91}$ \\

PolyCE-1 Loss & 
$78.81_{\pm 4.46}$ & 
$90.94_{\pm 0.51}$ & 
$91.71_{\pm 2.32}$ & 
$63.83_{\pm 6.26}$ \\

Tversky Loss & 
$79.69_{\pm 4.56}$ & 
$90.06_{\pm 0.68}$ & 
$91.41_{\pm 2.90}$ & 
$58.46_{\pm 6.77}$ \\

Focal Tversky Loss & 
$\mathbf{80.93_{\pm 4.21}}$ & 
$90.41_{\pm 0.38}$ & 
$91.97_{\pm 2.43}$ & 
$56.14_{\pm 6.58}$ \\

\rowcolor{black!5}{DropDice Loss} & 
$79.29_{\pm 4.68}$ & 
$90.21_{\pm 0.53}$ & 
$91.99_{\pm 2.45}$ & 
$70.49_{\pm 5.54}$ \\

\rowcolor{black!5}PolyDice-1 Loss & 
$80.73_{\pm 4.62}$ & 
$\mathbf{90.98_{\pm 0.63}}$ & 
$\mathbf{92.30_{\pm 2.22}}$ & 
$\mathbf{70.78_{\pm 5.20}}$ \\
\bottomrule[0.4mm]
\end{tabular}
\end{table*}
}

\def\tabletransunet{
\begin{table*}[t!]
\centering
\caption{Comparison of Dice Similarity Coefficient (\%) for TransUNet. Results are reported as the mean and standard deviation over 5-fold cross-validation, and higher values indicate better segmentation performance. The best score for each dataset is shown in bold, and shaded rows denote the proposed losses.}
\label{tab:trans_unet_results}
\begin{tabular}{lcccc}
\toprule[0.4mm]
\textbf{Method} & \textbf{CVC-ClinicDB} & \textbf{Kvasir-SEG} & \textbf{ACDC} & \textbf{Synapse} \\
\midrule
Dice Loss & 
$91.03_{\pm 4.29}$ & 
$93.90_{\pm 0.50}$ & 
$93.01_{\pm 0.78}$ & 
$78.73_{\pm 6.05}$ \\

Cross Entropy Loss & 
$90.73_{\pm 3.87}$ & 
$93.48_{\pm 0.39}$ & 
$91.97_{\pm 0.84}$ & 
$73.71_{\pm 6.01}$ \\

PolyCE-1 Loss & 
$90.67_{\pm 4.12}$ & 
$93.44_{\pm 0.50}$ & 
$92.45_{\pm 0.66}$ & 
$73.31_{\pm 8.05}$ \\

Tversky Loss & 
$\mathbf{91.35_{\pm 4.01}}$ & 
$93.11_{\pm 0.33}$ & 
$92.52_{\pm 0.66}$ & 
$52.34_{\pm 4.83}$ \\

Focal Tversky Loss & 
$91.02_{\pm 4.50}$ & 
$93.13_{\pm 0.48}$ & 
$92.82_{\pm 0.70}$ & 
$56.55_{\pm 6.76}$ \\

\rowcolor{black!5}DropDice Loss & 
$91.19_{\pm 4.27}$ & 
$\mathbf{93.92_{\pm 0.43}}$ & 
$93.03_{\pm 0.69}$ & 
$\mathbf{79.23_{\pm 6.18}}$ \\

\rowcolor{black!5}PolyDice-1 Loss & 
$90.97_{\pm 4.73}$ & 
$93.79_{\pm 0.40}$ & 
$\mathbf{93.20_{\pm 0.70}}$ & 
$79.16_{\pm 5.78}$ \\
\bottomrule[0.4mm]
\end{tabular}
\end{table*}
}

%% file: figs.tex
\def\overview{
\begin{figure}[t]
\centering
\includegraphics[width=1.0\linewidth]{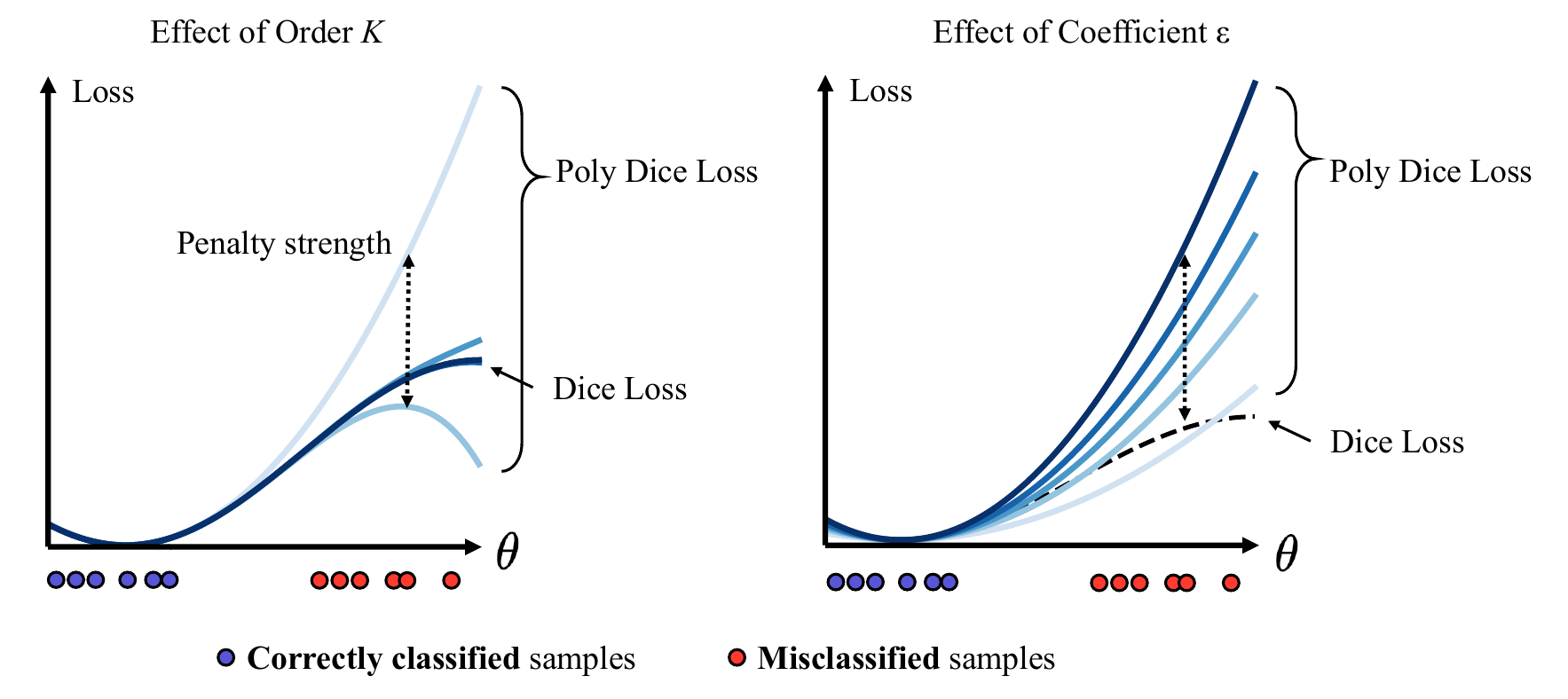}
\caption{Overview of the proposed Polynomial Dice Loss. Dice Loss is decomposed into a scale term and an angular alignment term, and the angular term is expanded as a polynomial by Taylor approximation. The truncation order $K$ in DropDice and the leading coefficient $\epsilon_1$ in PolyDice-1 control the penalty assigned to misaligned predictions.}
\label{fig:overview}
\end{figure}
}

\def\visdropandpoly{
\begin{figure}[t]
\centering
\begin{minipage}[b]{0.45\linewidth}{
\centering
\includegraphics[width=1.0\linewidth]{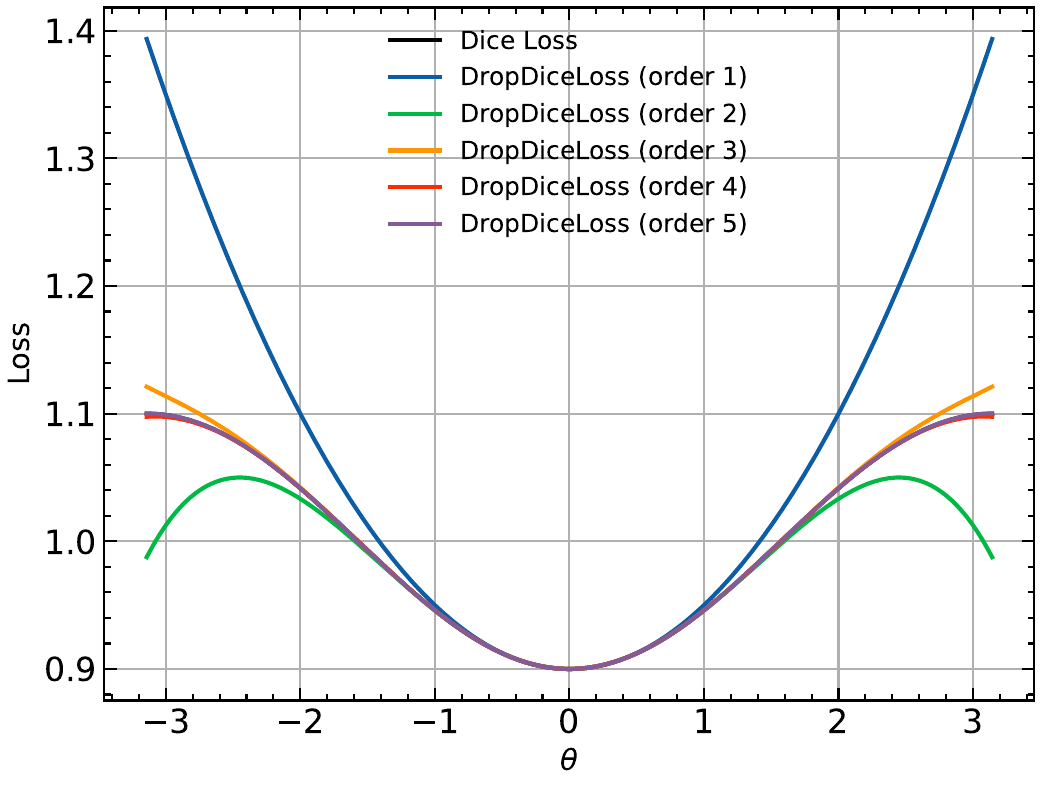}
\caption{DropDice Loss for $s=0.1$: smaller $K$ penalizes large angular errors more strongly, while larger $K$ approaches Dice Loss.}
\label{fig:dropdice}}
\end{minipage}
 \hspace{1.0cm}
\begin{minipage}[b]{0.45\linewidth}{
\centering
\includegraphics[width=1.0\linewidth]{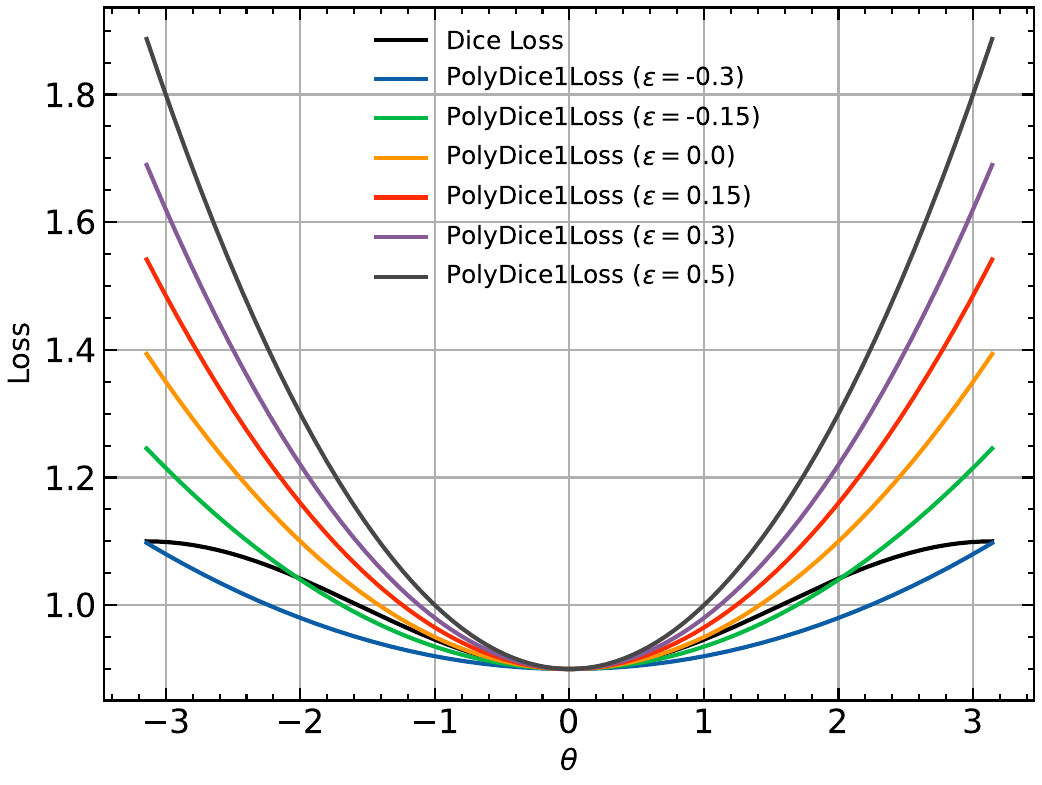}
\caption{PolyDice-1 Loss for $s=0.1$: larger $\epsilon_1$ strengthens the leading angular penalty, whereas smaller values are more conservative.}
\label{fig:poly1dice}}
\end{minipage}
\end{figure}
}

\newcommand{\colwidthdrop}{0.48\textwidth}
\def\expdropdice{
\begin{figure*}[t]
\centering
\begin{minipage}[b]{\colwidthdrop}
\centering
\includegraphics[width=1.0\linewidth]{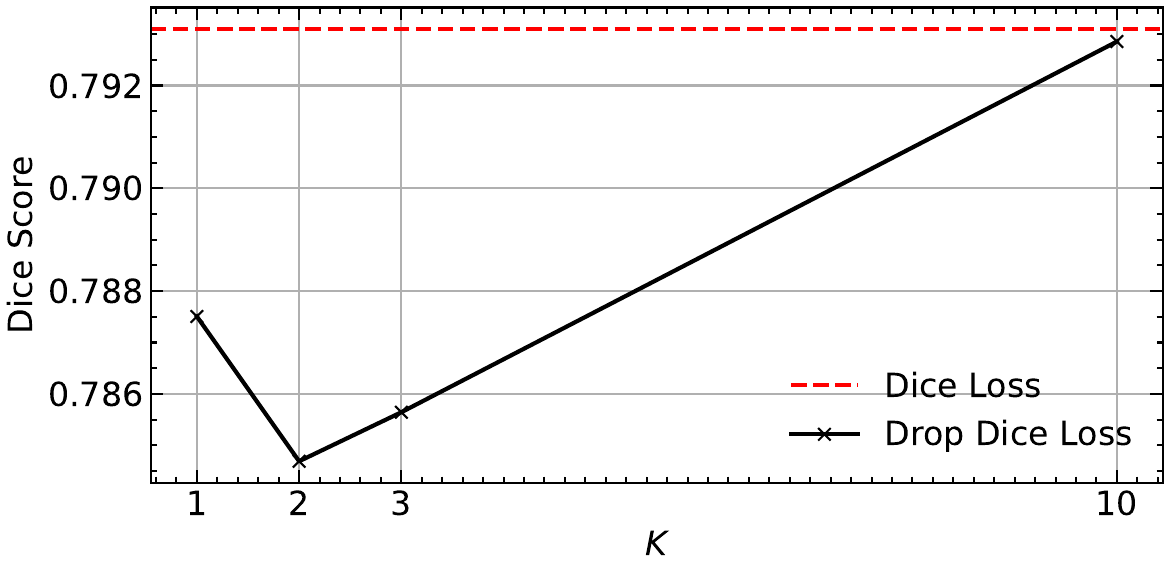}
\vspace{0.3em}
{\footnotesize CVC-ClinicDB}
\end{minipage}
\begin{minipage}[b]{\colwidthdrop}
\centering
\includegraphics[width=1.0\linewidth]{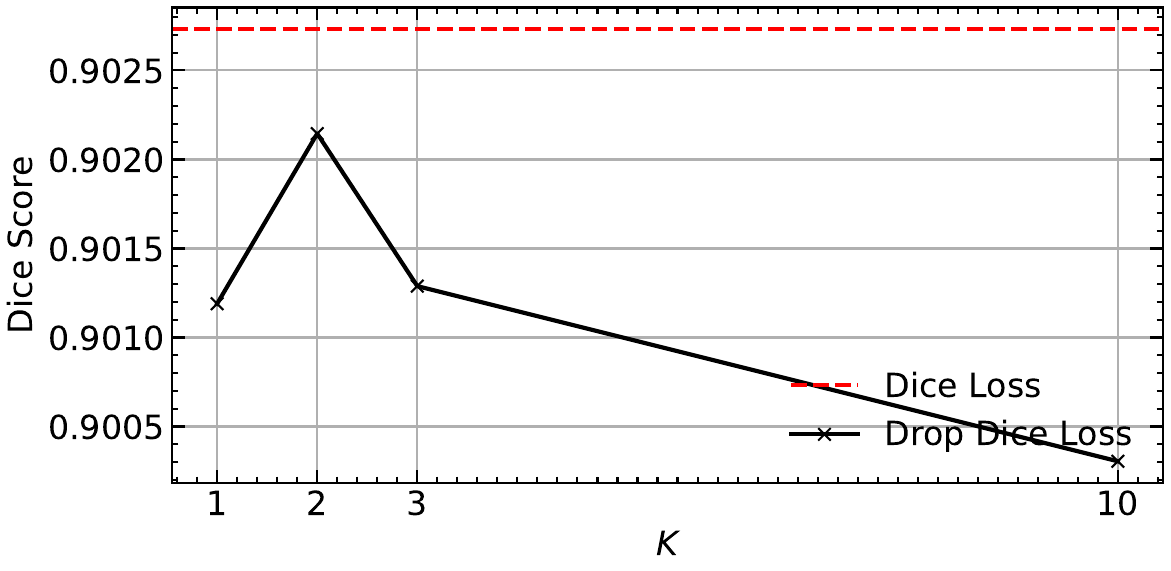}
\vspace{0.3em}
{\footnotesize Kvasir-SEG}
\end{minipage}
\begin{minipage}[b]{\colwidthdrop}
\centering
\includegraphics[width=1.0\linewidth]{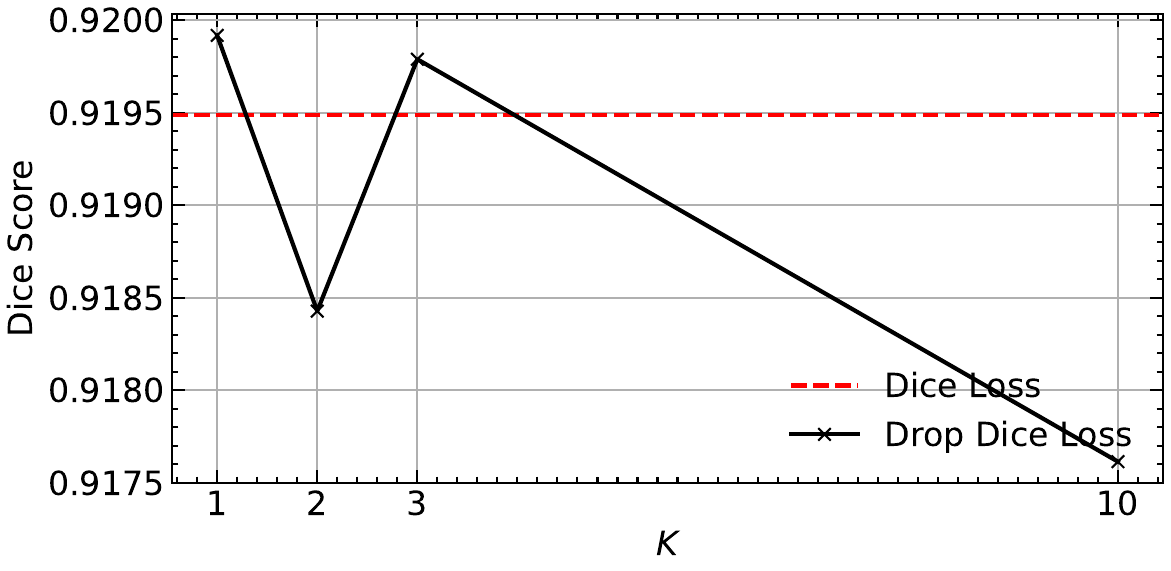}
\vspace{0.3em}
{\footnotesize ACDC}
\end{minipage}
\begin{minipage}[b]{\colwidthdrop}
\centering
\includegraphics[width=1.0\linewidth]{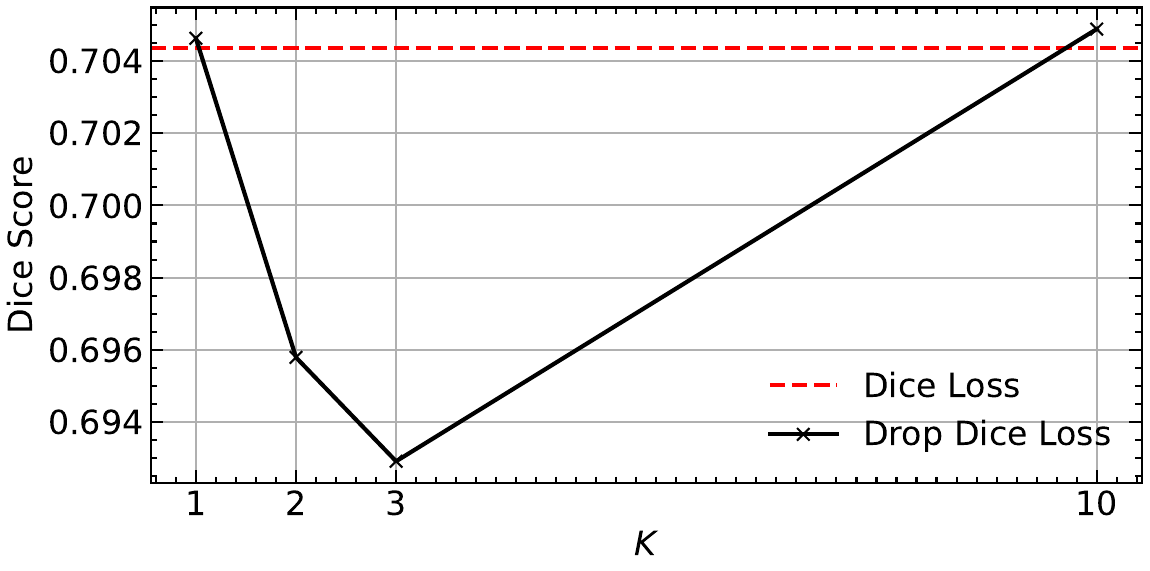}
\vspace{0.3em}
{\footnotesize Synapse}
\end{minipage}
\caption{Effect of the truncation order $K$ in $\ell^{(K)}_{\mathrm{DropDice}}$ for UNet. Dice scores are shown for four datasets, illustrating that the best order is dataset dependent: larger $K$ often approaches or improves upon Dice Loss, while smaller orders can be beneficial on some datasets.}
\label{fig:exp_drop_dice_k}
\end{figure*}
}

\newcommand{\colwidthpoly}{0.48\textwidth}
\def\exppolydice{
\begin{figure*}[t]
\centering
\begin{minipage}[b]{\colwidthpoly}
\centering
\includegraphics[width=1.0\linewidth]{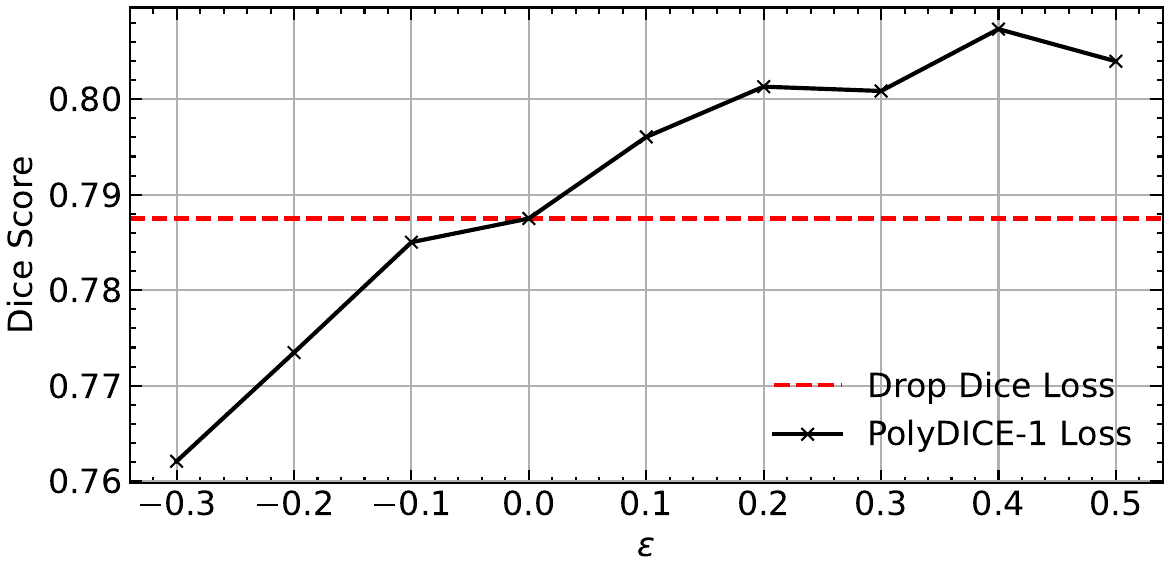}
\vspace{0.3em}
{\footnotesize CVC-ClinicDB}
\end{minipage}
\begin{minipage}[b]{\colwidthpoly}
\centering
\includegraphics[width=1.0\linewidth]{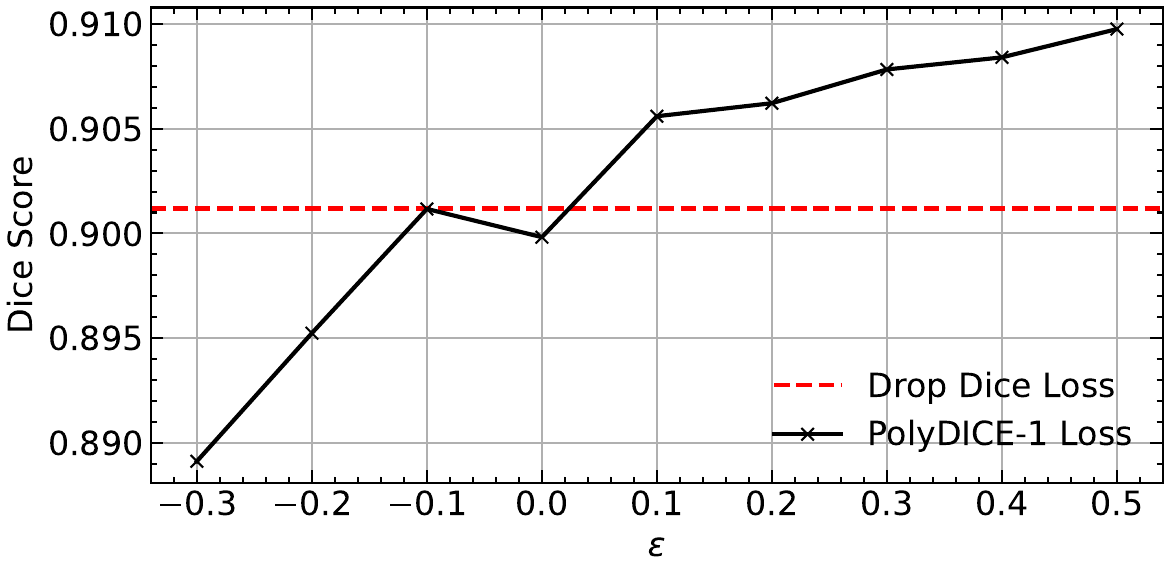}
\vspace{0.3em}
{\footnotesize Kvasir-SEG}
\end{minipage}
\begin{minipage}[b]{\colwidthpoly}
\centering
\includegraphics[width=1.0\linewidth]{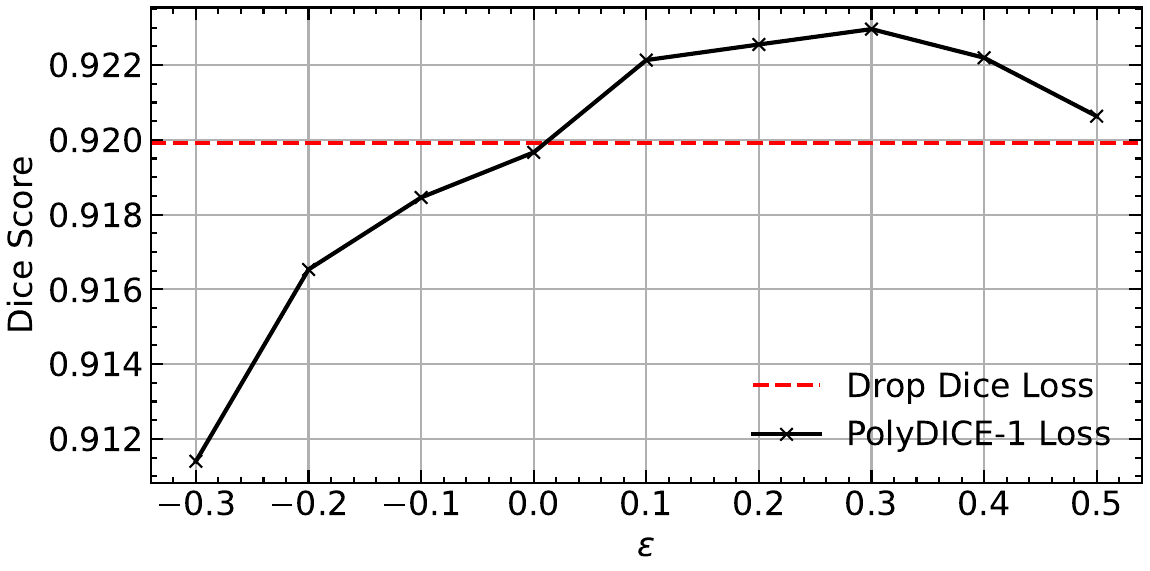}
\vspace{0.3em}
{\footnotesize ACDC}
\end{minipage}
\begin{minipage}[b]{\colwidthpoly}
\centering
\includegraphics[width=1.0\linewidth]{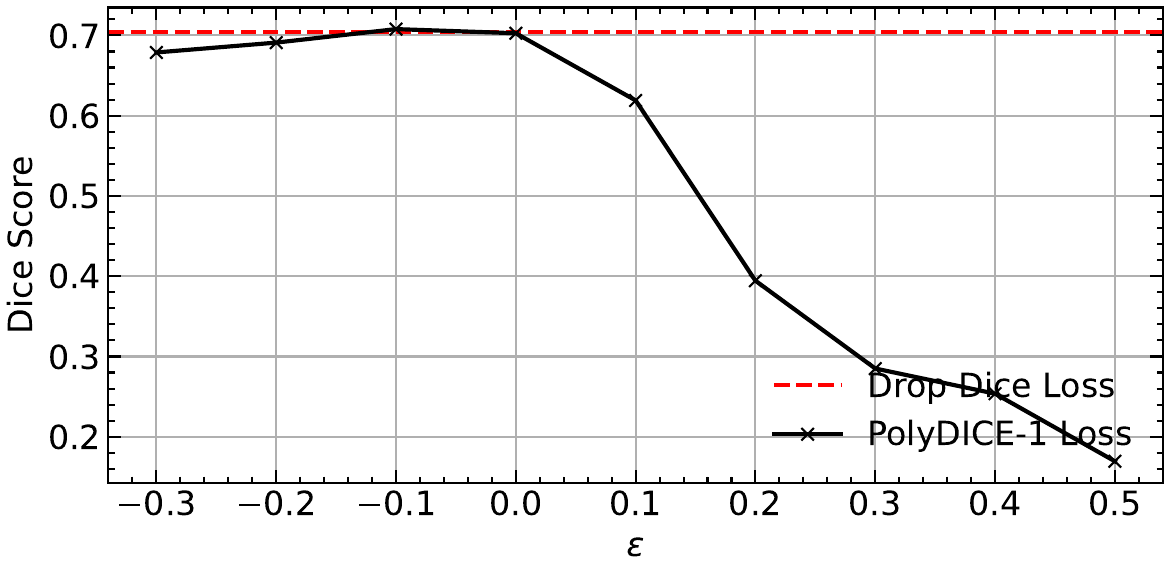}
\vspace{0.3em}
{\footnotesize Synapse}
\end{minipage}
\caption{Effect of the coefficient $\epsilon$ in $\ell_{\mathrm{PolyDice}\text{-}1}$ for UNet. We vary the single coefficient that controls the leading angular penalty and report Dice scores on each dataset, showing that validation-based tuning of $\epsilon$ affects performance and that the optimal value differs across datasets.}
\label{fig:exp_poly_dice_epsilon}
\end{figure*}
}

\newcommand{\colwidth}{0.15\textwidth}
\def\resultseg{
\begin{figure*}[t!]
    \centering
    \begin{minipage}[b]{\colwidth}
        \centering
        \includegraphics[width=1.0\linewidth]{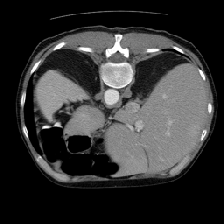}
    \end{minipage}
    \hfill
    \begin{minipage}[b]{\colwidth}
        \centering
        \includegraphics[width=1.0\linewidth]{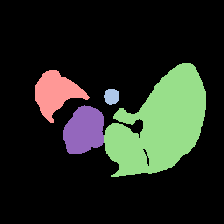}
    \end{minipage}
    \hfill
    \begin{minipage}[b]{\colwidth}
        \centering
        \includegraphics[width=1.0\linewidth]{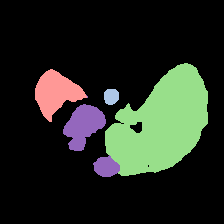}
    \end{minipage}
    \hfill
    \begin{minipage}[b]{\colwidth}
        \centering
        \includegraphics[width=1.0\linewidth]{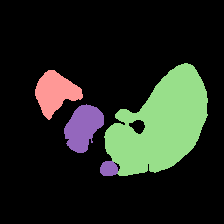}
    \end{minipage}
    \hfill
    \begin{minipage}[b]{\colwidth}
        \centering
        \includegraphics[width=1.0\linewidth]{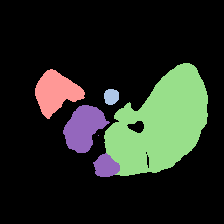}
    \end{minipage}
    \hfill
    \begin{minipage}[b]{\colwidth}
        \centering
        \includegraphics[width=1.0\linewidth]{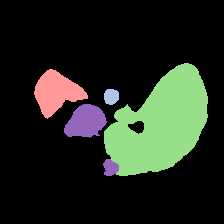}
    \end{minipage}

    \vspace{0.1cm}

    \begin{minipage}[b]{\colwidth}
        \centering
        \includegraphics[width=1.0\linewidth]{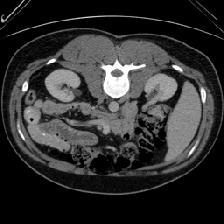}
    \end{minipage}
    \hfill
    \begin{minipage}[b]{\colwidth}
        \centering
        \includegraphics[width=1.0\linewidth]{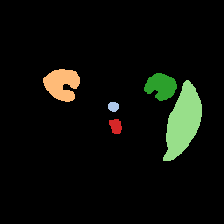}
    \end{minipage}
    \hfill
    \begin{minipage}[b]{\colwidth}
        \centering
        \includegraphics[width=1.0\linewidth]{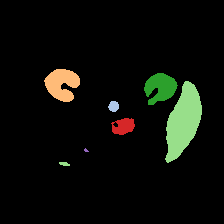}
    \end{minipage}
    \hfill
    \begin{minipage}[b]{\colwidth}
        \centering
        \includegraphics[width=1.0\linewidth]{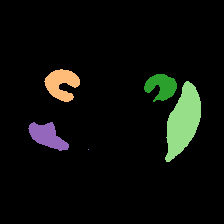}
    \end{minipage}
    \hfill
    \begin{minipage}[b]{\colwidth}
        \centering
        \includegraphics[width=1.0\linewidth]{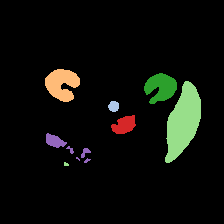}
    \end{minipage}
    \hfill
    \begin{minipage}[b]{\colwidth}
        \centering
        \includegraphics[width=1.0\linewidth]{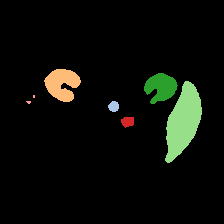}
    \end{minipage}

    \vspace{0.1cm}
    
    \begin{minipage}[b]{\colwidth}
        \centering
        \includegraphics[width=1.0\linewidth]{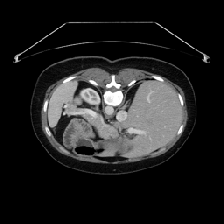}
        \vspace{0.3em}
        {\scriptsize Input}
    \end{minipage}
    \hfill
    \begin{minipage}[b]{\colwidth}
        \centering
        \includegraphics[width=1.0\linewidth]{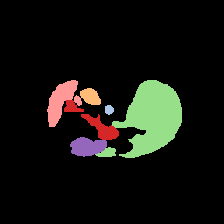}
        \vspace{0.3em}
        {\scriptsize GT}
    \end{minipage}
    \hfill
    \begin{minipage}[b]{\colwidth}
        \centering
        \includegraphics[width=1.0\linewidth]{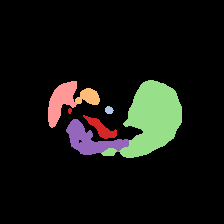}
        \vspace{0.3em}
        {\scriptsize Dice}
    \end{minipage}
    \hfill
    \begin{minipage}[b]{\colwidth}
        \centering
        \includegraphics[width=1.0\linewidth]{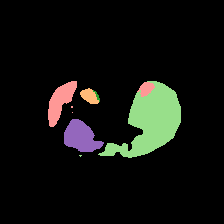}
        \vspace{0.3em}
        {\scriptsize Tversky}
    \end{minipage}
    \hfill
    \begin{minipage}[b]{\colwidth}
        \centering
        \includegraphics[width=1.0\linewidth]{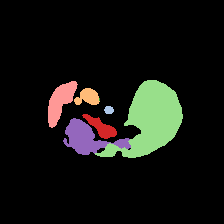}
        \vspace{0.3em}
        {\scriptsize DropDice}
    \end{minipage}
    \hfill
    \begin{minipage}[b]{\colwidth}
        \centering
        \includegraphics[width=1.0\linewidth]{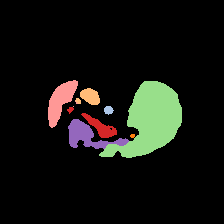}
        \vspace{0.3em}
        {\scriptsize PolyDice-1}
    \end{minipage}

    \caption{Qualitative segmentation results on the Synapse dataset using TransUNet. Each row shows one sample, with columns corresponding to the input image, ground truth, Dice Loss, Tversky Loss, DropDice Loss, and PolyDice-1 Loss. In these examples, the proposed losses preserve some organ regions more clearly than the baselines, consistent with the observed quantitative trends on Synapse.}
    \label{fig:results_seg}
\end{figure*}
}